%% file: relign.tex
\documentclass[letterpaper, 10pt, conference]{ieeeconf}

\IEEEoverridecommandlockouts                              
\usepackage[pdftex]{graphics} 
\usepackage{epsfig} 
\usepackage{mathptmx} 
\usepackage{times} 
\usepackage{amsmath} 
\usepackage{amssymb}  
\usepackage{subcaption}
\usepackage{url}
\usepackage{hyperref} 

\usepackage{tikz}
\usepackage{tkz-berge}

\usetikzlibrary{patterns}
\usetikzlibrary{positioning}
\usetikzlibrary{patterns}
\usetikzlibrary{matrix}
\usetikzlibrary{shapes}
\usetikzlibrary{decorations.pathreplacing}
\usetikzlibrary{graphs, graphs.standard}
\usetikzlibrary{fadings}

\usepackage{tikz-3dplot}

\def\eps{{\epsilon}}
\DeclareMathOperator*{\argmin}{arg\,min}
\newcommand{\ring}[1]{\ensuremath{\mathbb{#1}}}
\newcommand\RR{\ring{R}}
\newcommand\NN{\ring{N}}
\newcommand{\range}{S}
\newcommand{\Woff}{W_{\mathrm{off}}}
\newcommand{\Wdist}{W_{\mathrm{dist}}}
\newcommand{\package}{\emph{relign}}
\newcommand{\WL}{W_{\mathrm{L}}}

\title{\LARGE \bf
Active Alignments of Lens Systems with Reinforcement Learning*
}

\author{Matthias Burkhardt, Tobias Schmähling, Pascal Stegmann, Michael
    Layh, and Tobias Windisch
\thanks{*This work was funded by the German Federal Ministry of Education and Research (BMBF) under
    grant number 13FH605KX2.
This work has been submitted to the IEEE for possible publication. Copyright may be transferred without notice, after which this version may no longer be accessible.}
\thanks{All authors are with the Institute for Machine Vision, 
        University of Applied Sciences Kempten, 87435 Kempten, Germany.
     (email: \{matthias.burkhardt, tobias.schmaehling, pascal.stegmann, michael.layh, tobias.windisch\}@hs-kempten.de).
    }
}

\begin{document}

\maketitle
\thispagestyle{empty}
\pagestyle{empty}

\begin{abstract}
	Aligning a lens system relative to an imager is a critical challenge
	in camera manufacturing.
	While optimal alignment can be mathematically computed under ideal conditions, real-world
	deviations caused by manufacturing tolerances often render this approach impractical. Measuring
	these tolerances
	can be costly or even infeasible, and neglecting them may result in suboptimal alignments.
	We propose a reinforcement learning (RL) approach that learns
	exclusively in the pixel space of the sensor output, eliminating the
	need to develop expert-designed alignment concepts.
	We conduct an extensive benchmark study
	and show that our approach surpasses other methods in
	speed, precision, and robustness.
	We further introduce \package{}, a realistic, freely
	explorable, open-source
	simulation utilizing physically based rendering that models optical systems
	with non-deterministic manufacturing tolerances and noise in robotic
	alignment movement. It provides an interface to popular machine learning
	frameworks, enabling seamless experimentation and development.
	Our work highlights the potential of RL in a manufacturing environment to enhance
	efficiency of optical alignments while minimizing the need for manual intervention.
\end{abstract}

\section{Introduction}
The assembly of optical devices requires precise positioning when joining their individual
components. This requirement is essential in a wide range of products, including cameras in mobile
phones, fiber optics, aerial and medical imaging and optical projection systems for
microlithography~\cite{yoder2008mounting}. One particularly sensitive process is an alignment,
where two components must be precisely positioned relative to each other to achieve high precision.
A prominent example is the positioning of a lens system relative to an
imager~\cite{alignment_strategies}. The compound product must be assembled in a way that the
optical performance is maximized. While high-cost lenses are often designed to ease the alignment
with an imager, achieving optimal alignment with low-cost components presents a significant
challenge. All of these components typically offer many degrees of freedom, each influencing
multiple performance metrics in complex and interdependent ways. Often, it is unclear how the
position has to be modified in order to reach a performance satisfying predefined quality
constraints. Additionally, variations of the components make the relations between the position and
the optical performance diffuse and noisy. These challenges have been widely studied in the
literature~\cite{RobotAssemblyLargeScaleObjects, PoseEstimationLargeScaleObjects,
	FastAlignmentPhotonicsDevice,offline_alignment}.

\begin{figure}[t]
	\centering
	\scalebox{0.8}{
		\input{./figures/overview.tex}
	}
	\caption{Schematic presentation of a single alignment step, where a lens
		system consisting of four single lenses (left) has to be positioned
		relative to an optical sensor (right).}
	\label{fig:OpticalAlignments}
\end{figure}

The classic way to deal with such problems involves extensive scans during the alignment of each
optical system individually, where any degrees of freedom are varied and evaluated separately. To
make those scans robust against manufacturing tolerances within the components, many possible
positions are scanned by a structured walk through the alignment space, for example along
coordinate axes. Often, these algorithms solely rely on hand-crafted features obtained from the
high-dimensional sensor output, where sensor and movement noise make it hard to conduct
deterministic algorithms. Speeding up optical alignments has thus been a fruitful application of
machine learning methods in the past, see~\cite{rakhmatulin2024review} for a review. Some
approaches predict next alignment moves from misaligned settings in a supervised
fashion~\cite{supervised_alignments,ml_for_off_axis_systems}. A detailed study for Fast-Axis
Collimating Lenses can be found in~\cite{ModelBasedAlignment}. In their basic form, however, active
alignment problems are no supervised learning problems. This is due to the fact that first,
symmetries and offsets in the optical layout make it hard---or even impossible---to set up a
supervised dataset from the sensor observation to the optimal sensor image. Second, training models
via supervised learning cannot account to minimize the number of alignment steps. For instance,
sometimes a step into the wrong direction has to be taken in order to explore symmetries in the
robotic movements. Thus, more naturally, optical alignments are modeled as an RL problem which
canonically allows training models to find short trajectories to optimal positions. RL algorithms
have demonstrated the ability to learn complex relationships for various challenging
tasks~\cite{alphago,dreamer,alphafold}. There has also been plenty of research using RL for process
control in manufacturing (see~\cite{rl_for_process_control} and references therein). For the
alignment of laser optics or interferometers, RL has already been applied successfully as
demonstrated in~\cite{rl_for_laser_optics} or~\cite{sorokin2020interferobot}, respectively.
Particularly when applied to real systems, RL comes with its own intrinsic
challenges~\cite{rl_challenges}, like sparse and delayed rewards~\cite{delayed_reward}, data
inefficiencies~\cite{rl_efficiency}, and reproducibility issues~\cite{rl_reproducability},
rendering high need for research when applied to new tasks.

In this study, we formulate optical alignment problems as an RL task, where optimal robotic
alignment movements are learned solely in the pixel space from a high-dimensional sensor
observation (see Figure~\ref{fig:OpticalAlignments}). The alignment goal is reached when the
difference between the observed image and a given reference pattern falls below a predefined
threshold. We study different reward functions to motivate RL agents to find optimal alignment
positions in as few steps as possible. To the best of our knowledge, our work is the first that
treats an optical alignment task as a \emph{Partially Observable Markov Decision Process} (POMDP).
More specifically, our main contributions are:

\begin{itemize}
	\item We formulate general active alignment problems as POMDP tangible by RL algorithms. To the best of
	      our knowledge, this is the first work in doing so.
	\item We introduce the modular Python framework~\package{}\footnote{Code
        under~\url{https://github.com/hs-kempten/relign}.} for simulating active alignment scenarios using the physically based rendering
	      framework Mitsuba~\cite{Mitsuba3}. Its interface is compatible with the Gymnasium
	      API~\cite{gymnasium} and allows effective benchmarking of state-of-the-art methods on
	      representative alignment tasks.
	\item We show that RL algorithms can solve real-world-inspired alignment problems more efficiently than
	      other methods based on machine learning, even under presence of manufacturing tolerances and noise
	      in robotic alignment movements while maintaining high accuracy and low inference time.
\end{itemize}

Finding accurate and comparable benchmarks for active alignments is hard, not only due to privacy
constraints of companies maintaining them as part of their core know-how, but also due to the fact
that optical products are very diverse and have different requirements. Here, our framework
provides a first step towards a common benchmarking environment for active alignments. Moreover, we
argue that using RL for alignments not only has the potential to significantly speed up the optical
alignment process but also renders the need for designing hand-crafted features obsolete.

\section{Active Alignments of Lens Systems}\label{s:active-alignments}

Automated robotic alignment systems often rely on high-precision motion platforms such as hexapods
or piezo stages, which provide movement in all six degrees of freedom. These mechanisms enable
sub-micrometer adjustments, allowing the robot to finely correct tilt, shift, and focus errors
until the desired alignment is reached. In this section, we formalize the active alignment problem
and introduce the main challenges faced during the alignment process.

\subsection{Problem Formulation}

An optical alignment process joins multiple components to maximize optical performance. In this
work, the component to be aligned is a lens system consisting of a fixed number of single lenses
$L=(L_1,L_2, L_3, L_4)$ that need to be positioned relative to an optical sensor. This situation is
typical when manufacturing cameras. The goal of an alignment is to move $L$ to a position $s=(x, y,
	z, R_x, R_y, R_z)\in\mathbb{R}^{6}$ relative to a sensor, typically with an automated robotic
alignment system, such that the optical performance is maximal.

The main challenge is that in each alignment, the process has to adjust to new conditions, mainly
due to variances within $L$ and when gripping $L$. The first type of variance is a randomized
offset arising from the initial placement of $L$ in the alignment station. This can be modeled by a
randomized starting position $\Woff\in\RR^6$ representing a random translation and rotation offset.
When connecting to the robotic alignment system, the movement is typically not optimal, meaning
that when a movement $a\in\RR^6$ in a state $s$ is executed, the new state is not $s+a$ but the
slightly distorted state $s+\Wdist\cdot a$ with $\Wdist\in\RR^{6\times 6}$. Moreover, the new state
can be clipped, for instance because a boundary condition is met. We refer to
Section~\ref{s:change-position} for more details on how $\Wdist$ is constructed and how the
boundary behavior is modeled. Another type of variance comes from the production of $L$ itself,
which cannot be changed by the process during the alignment. That is, each single lens $L_i$ in $L$
has an individual tilt and position offset in comparison to the \emph{ideal} lens system. We denote
the offsets of each lens within $L$ by $\WL\in\RR^{k\times 6}$. Other variances not considered in
this work are dispersions in the geometries of the single lenses, like their curvature. Here, the
variances $W=(\Woff, \Wdist, \WL)$ characterize an alignment completely and we assume that these
are sampled from an unknown distribution $W\sim \rho$. We further assume that the variances $W$ are
latent to the alignment process, i.e., cannot be measured while aligning. As a result, it is not
possible to directly compute the position where optical performance is optimal using the physical
equations of e.g. geometrical optics.

\subsection{Performance Measurements}

\begin{figure}
	\includegraphics[width=0.45\textwidth]{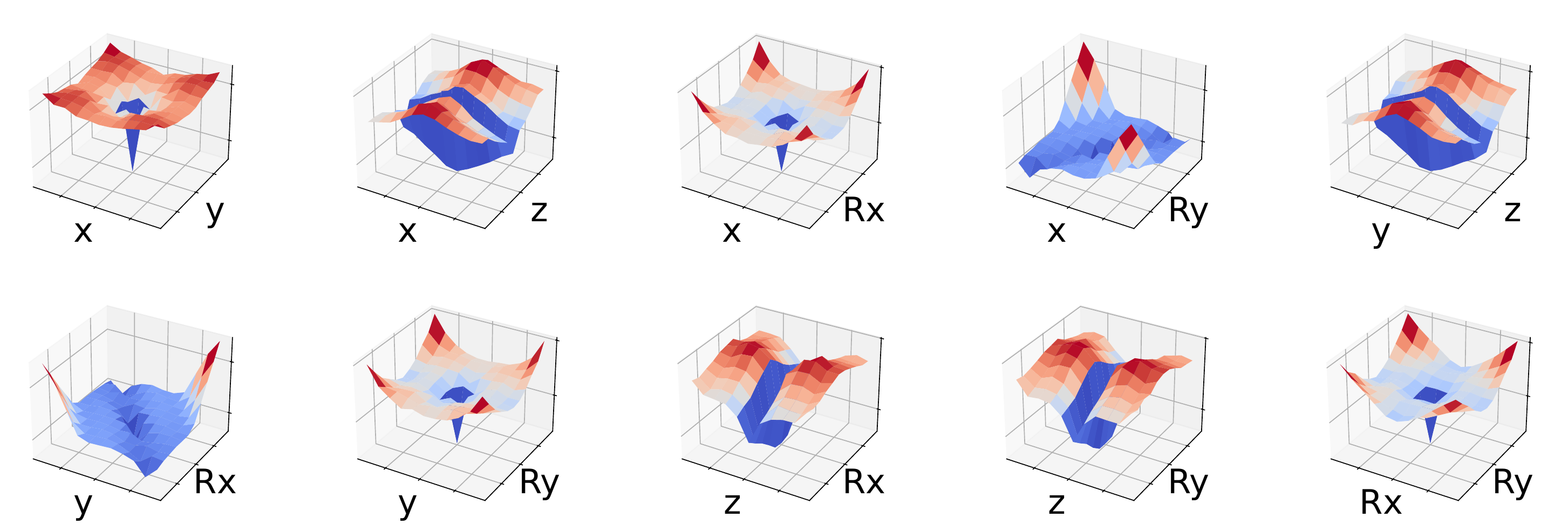}
	\caption{Visualization of the two-dimensional projection $(s_i, s_j)\mapsto \|O(s^*-s_ie_i-s_je_j,
			W^*)-O^*\|$ for each tuple
		$\{i,j\}\subset\{x,y,z,R_x,R_y\}$, that is, all except two dimensions are fixed to the
		optimal values.}\label{fig:optimality}
\end{figure}

To quantify the optical performance at a given state~$s$, collimated light is sent through $L$ and
measured at the sensor with width $w$ and height $h$, yielding a high-dimensional image $O(s,
	W)\in\RR^{w\times h}$ of light intensities. The sensor output $O(s, W)$ is noisy, and retrieving
$O(s, W)$ multiple times for the same position $s$ always yields slightly different observations.
We call this noise \emph{sensor variance}. In many industrial applications, hand-crafted scalar
features are extracted from $O(s, W)$, typically involving the optical transfer functions, for
which quality bounds have to be reached during the alignment. Here, however, we study problems
where a generic \emph{reference output} $O^*$ independent of $W$ is given such that the alignment
task is the identification of a state $s_W$ where the sensor output matches the reference, i.e.,
$O(s_W, W)\approx O^*$. The reference pattern can be considered as $O^*=\mathbb{E}[O_{W}(s_W)]$,
that is the sensor output without noise $W^*=0$ in the positioning of $L$ and its lenses at the
respective optimal position $s_W$. In practice, the choice of the light field used for creating a
test pattern at sensor level is dictated by the specific application requirements, such as whether
the camera needs to achieve sharp focus in the near or far field. A common example is the Siemens
star~\cite{siemens_star}, which is widely used for evaluating optical performance of digital
cameras~\cite{iso}. Numerous methods exist to measure the quality of an optical image based on
projected patterns such as a pattern of Siemens stars and slanted edges as described in
ISO~12233~\cite{iso}. Given a reference pattern, the optimization problem
\begin{equation}\label{equ:optimality}
	\argmin_{s\in S}\|O(s, W)-O^*\|
\end{equation}
has to be solved for a given situation $W\sim\rho$, where $S\subset\RR^6$ denotes
the set of allowed positions. Here, $\|\cdot\|$ denotes a vector norm, like the
Euclidean distance, and we interpret the input matrices as vectors.
Figure~\ref{fig:optimality}
shows some 2D-projections of this optimization problem
for the alignment situations introduced in Section~\ref{s:experiment}.
Typically, a threshold
$\theta\in\RR_{\ge 0}$ is
given such that any $s$ that satisfies
$\|O(s, W)-O^*\|\le \theta$ is considered as \emph{optimal}.
However, as $s$ can only compensate positional offsets of
$L$ given by $\Woff$ under distortions given by $\Wdist$ and not
the variances $\WL$ inside $L$, some lens systems
may have $\|O(s, W)-O^*\|>\theta$ for all positions $s$.

\subsection{Alignment Algorithms}\label{s:alignment_algo}

As the variances $W$ are unobservable by the alignment process, most active alignment algorithms
explore the state space~$S$ iteratively by selecting states and searching for directions where the
score $\|O(s, W)-O^*\|$ attains its minimum. Once positioned at a new state, a new sensor signal is
observed. An \emph{alignment algorithm} computes a series of subsequent actions $a_1,\ldots,a_n$,
step by step, starting from a randomized initial position $s_0=\Woff$. Each action is an element of
a set of allowed actions $A\subset\RR^6$ and generates a sequence of states
$s_i=s_0+\Wdist(a_1+\ldots+a_i)$ such that $s_n$ is optimal, i.e., $\|O(s_n, W)-O^*\|\le\theta$.
The computation of~$a_i$ must be based on a subset of the $i-1$ many images $O(s_1, W),\ldots,
	O(s_{i-1}, W)$ obtained so far. In Section~\ref{s:environment}, we state the alignment algorithms
used in our benchmark study.

\section{Alignments as a POMDP}\label{s:pomdp}

In this section, we describe how RL algorithms can be used as alignment algorithms as defined in
Section~\ref{s:alignment_algo}. Specifically, we consider active alignments as an \emph{episodic}
POMDP, where each episode is the alignment of a given lens system $L$. Here, a state $(s, W)$ is
decomposed in a position $s\in\range$ and variances $W=(\Woff, \Wdist, \WL)$ as defined in
Section~\ref{s:active-alignments} and the set $A$ denotes the set of actions. Selecting $a$ at $(s,
	W)$ results in the new state $(s', W)$ with $s'=s+\Wdist\cdot a$, yielding a reward $R(s, s')$. As
of the state representing the variances stays constant through an episode, this MDP is also an
\emph{contextual} MDP in the sense of~\cite[Section~3]{contextual_mdp}. The state $(s, W)$ cannot
be directly observed.

Instead, only the high-dimensional sensor output $O(s, W)$ is given, which is controlled by a
conditional probability density function depending on $s$. We explain in
Section~\ref{s:sensor_sampling} how an image $O(s, W)$ is sampled from the sensor at a given state
$s$. For given $W$, an episode ends once a \emph{terminal state}, that is an optimal state from
$\{s\in S: \|O(s, W)-O^*\|\le\theta\}$, or an upper limit of steps $T\in\NN$ is reached. The goal
is to find a \emph{policy}~$\pi$ that maps observations to actions in a way that maximizes the
accumulated observed reward. More formally, given the observation $O(s, W)$ for $s$, the next state
is $s+\Wdist\cdot\pi(O(s, W))$. Starting from an initial state $s_0$, this combined dynamics of
sampled observations from the sensor and generated actions by the policy $\pi$ yields a trajectory
$(s_0,\ldots, s_k)$ of subsequent states with $s_{i+1}=s_i+\Wdist\cdot\pi(O(s_i, W))$ where the
last state $s_k$ is either optimal or $k=T$ holds. The goal is to find a policy such that
\begin{equation}\label{equ:rl_goal}
	\mathbb{E}_{W\sim\rho,(s_0,\ldots,
		s_k)\sim\pi}\left[\sum_{i=0}^{k-1}
		\gamma^i R(s_i, s_{i+1})\right]
\end{equation}
is maximized, where $\gamma\in(0,1)$ is the \emph{discount factor} given to
trade-off rewards in early and late states. Note that in general the reward
also depends on the action taken, which is not required in this work and hence
omitted.

To train effective agents for an alignment task, the reward function $R$ has to be designed in a
way that the optimization task from Equation~(\ref{equ:rl_goal}) is solved in a minimal number of
steps. During training time of a optical system $L$ with variances $W$, agents can be provided with
dense and insightful reward signals, for instance containing information of the distance to the
desired optimum $s_W$, which need to be known beforehand. An ideal reward signal would be
$-\|s-s_W\|$ which, however, requires the knowledge of $s_W$. In our simulation and during train
time, this could be computed, for instance using a conventional alignment algorithm, but this would
slow down training time as this needs to be done for each training episode anew. Instead, we use an
indirect signal which gives the distance to $s^*$, i.e. the optimal position if $W=0$:
$$R_{\text{dist}}(s)=-\|s-s^*\|.$$ However, $s_W$ may differ significantly from $s^*$, especially
in degenerated situations where $W$ is large. Another reward signal could be to use the reference
pattern $O^*$, which is known beforehand, like $$R_{\text{pattern}}(s, W)=-\|O(s, W)-O^*\|.$$ Yet,
even at the optimal position, $O(s_W, W)$ may differ significantly from $O^*$ and we found that for
complex patterns, the many local minima of $R_{\text{pattern}}$ make it hard for agents to learn
good policies from that reward signal alone. Thus, we considered a combination of both rewards
signals $$R(s, W)=
	\begin{cases}
		R_{\text{pattern}}(s, W) + R_{\text{dist}}(s), \quad & \text{if } \|s-s^*\|\le C \\
		R_{\text{pattern}}(s, W),                            & \text{otherwise}
	\end{cases}.
$$

This has the benefit that far away from $s^*$, and hence $s_W$, direction to the optimal region is
enforced. Once being sufficiently close to $s^*$, the agents are purely guided by the pattern
distance of the lens system at hand.

\section{Simulating Alignments}

One main challenge in simulating realistic optical alignments is to accurately model how a sensor
measures light emitted from a source and propagated through optical lenses. Here, we use Mitsuba3,
a physically based rendering engine for forward and inverse light-transport simulation. This not
only allows calculating light intensities $O(s, W)$ measured at the sensor (see
Section~\ref{s:sensor_sampling}) for a concrete position $s$, but also changing $s$ dynamically
(see Section~\ref{s:change-position}). All Mitsuba scenes consist of a sensor, a lens system with
four single lenses, and a light source. As one of our contribution is the open-source framework
\package{}, we put high engineering effort in providing a realistic simulation environment and
keeping the sim-to-real gap small.

\subsection{Sensor Outputs}\label{s:sensor_sampling}

The Mitsuba scene emulates collimated light from a rectangular area emitter source and a specific
reference pattern as binary bitmap. The light source is so far away that it can be considered
infinity. Starting at the sensor, Mitsuba traces light rays backwards that pass through the lenses.
We use an irradiance meter as sensor, which measures the incident power per unit area over a
predefined shape. In our setup, the sensor shape is $2 \times 2$ in Mitsuba space coordinates. To
solve the high-dimensional problem of rendering $O(s, W)$ numerically, Mitsuba employs Monte Carlo
integration, drawing $s_c$ many samples from a uniform distribution. As a consequence, the sensor
output $O(s, W)$ is different when rendering the same position $s$ with different seeds. This leads
to the probability density function of the POMDP as described in Section~\ref{s:pomdp}. An example
of the resulting sensor output is shown in Figure~\ref{fig:OpticalAlignments}. The measurements are
interpreted as a grayscale image and stored without any post-processing.

\subsection{Position Changes}\label{s:change-position}

Without loss of generality, we assume that the set of all possible states is the unit interval
$[0,1]^6$. We place our lens system $L$ in a way that $s^*:=(0.5,\ldots,0.5)\in\RR^6$ is the
optimal position. Upon initialization, we sample $\WL$ from a normal distribution and reposition
the single lenses within $L$ accordingly. Afterward, a starting vector $\Woff \in [0,1]^6$ is
sampled uniformly which defines the initial state $s_0:=\Woff$. The movement distortion matrix is
constructed as $\Wdist=I_6+\eps\in\RR^{6\times 6}$ where each coordinate in $\eps\in\RR^{6\times
		6}$ is sampled from a normal distribution. The positioning of the lens system can be varied by
setting an action $a\in\RR^6$ leading to an update $s'=s+\Wdist\cdot a$. To ensure that $s'$ stays
within $[0,1]^6$, each coordinate of $s+\Wdist\cdot a$ is clipped into $[0,1]$ before updating the
Mitsuba scene. Subsequently, the sensor output $O(s', W)$ is generated as described in
Section~\ref{s:sensor_sampling}.

\subsection{Generating the Reference Pattern}

To decide whether a state $s$ is in the optimality condition, the observation $O(s, W)$ has to be
compared to a reference pattern $O^*$. This reference pattern only depends on the optical layout of
the lens system, not on the noise level of $W$. To generate $W$ in our synthetic setting, we use a
perfectly aligned lens system where each of the single lenses are perfectly aligned as well, i.e.,
$W^*:=0$. We then sample $1.000$ observations from $O_{W^*}(s^*)$ and compute the pixel-wise mean
image as an approximation for $O^*=\mathbb{E}[O_{W^*}(s^*)]$.

\section{Measuring the sim-to-real gap}\label{s:optical_layout}

To emulate realistic manufacturing scenarios, we validate our simulation \package{} against an
established optical design software and images from a real-world alignment station. For all the
results presented in Section~\ref{s:experiment}, we used the following hand-designed lens system
was used.

\subsection{Comparison to Optical Design Software}
\begin{figure}
	\includegraphics[width=0.45\textwidth]{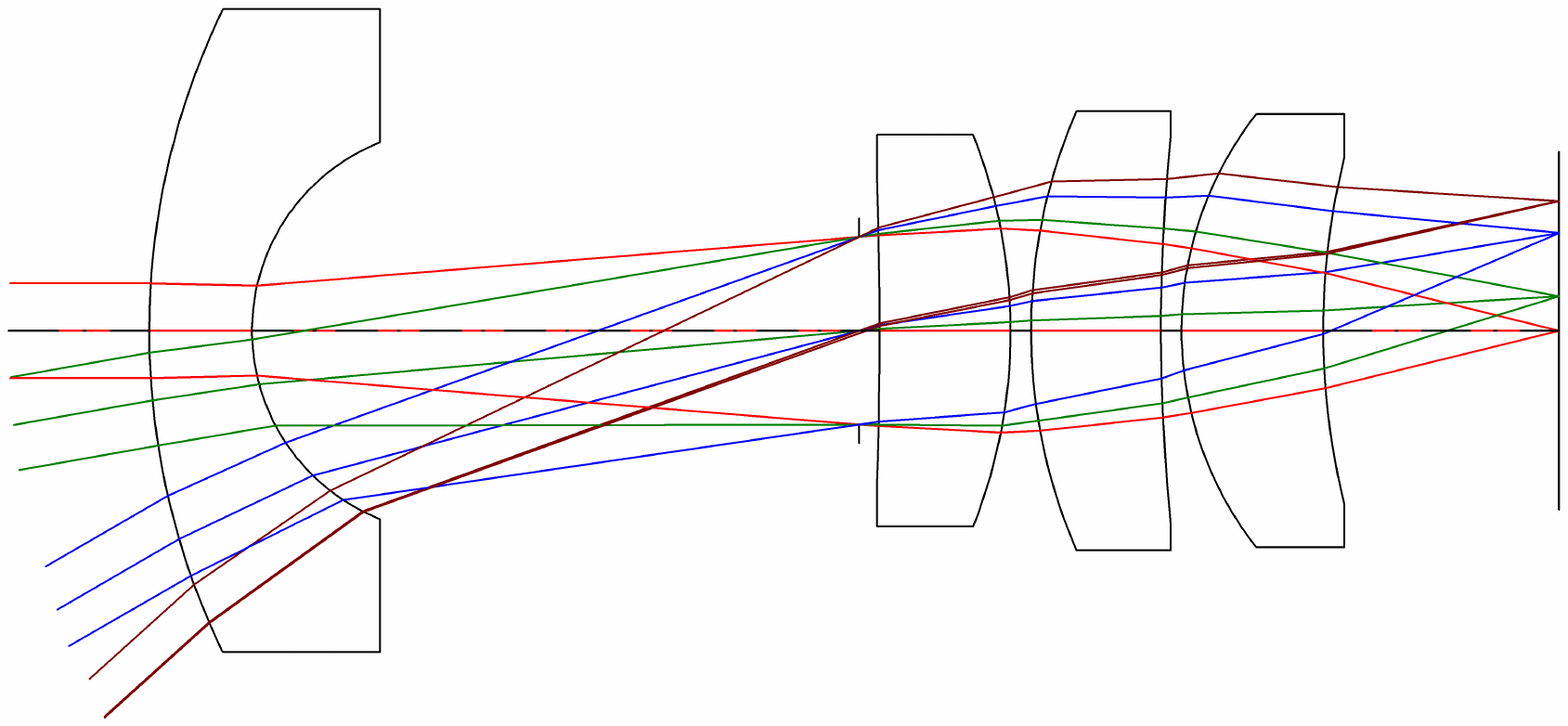}
	\caption{Optical diagram of lens system including light rays in Code V.}\label{fig:codev_lens}
\end{figure}

To assess the applicability of the proposed workflow to a more realistic optical system, we
designed a lens system consisting of four lenses in Code V~\cite{Synopsys_CodeV_2024} for
monochromatic operation at 589 nm. The system is optimized for imaging objects at infinity onto the
sensor plane and exhibits an effective focal length (EFL) of 2.895 mm, an f-number of 2.1, a field
of view (FOV) of 84.1°, and a half exit-pupil angle of 13.45°. The lens system has a mechanical
length of 20.59 mm and a maximum diameter of 9.8 mm. The design introduces a minimal distortion of
–27.2\% and throughout the entire field of view, the relative illumination does not fall below
60\%. Figure~\ref{fig:codev_lens} shows an optical diagram of the lens system including light rays
of different wave lengths.

For integration into \package{}, each lens has to be exported individually from Code V and
converted into polygonal meshes using any CAD tool (e.g., SolidWorks). Exporting the elements
separately, rather than the lens system as a whole, ensures that the components remain
independently addressable within the simulation environment. To enforce the defined field of view,
we introduced an additional mechanical mount and an aperture stop directly behind the first lens.

We modeled illumination by a rectangular area emitter providing diffuse emission with arbitrary
bitmap textures. The emitter dimensions and distance were chosen such that the system operates
within the optical field of view and the incident rays approximate collimated illumination,
consistent with the design specifications.

\begin{figure}
	\centering
	\subcaptionbox{\package{} (ours)}{\includegraphics[width=0.15\textwidth]{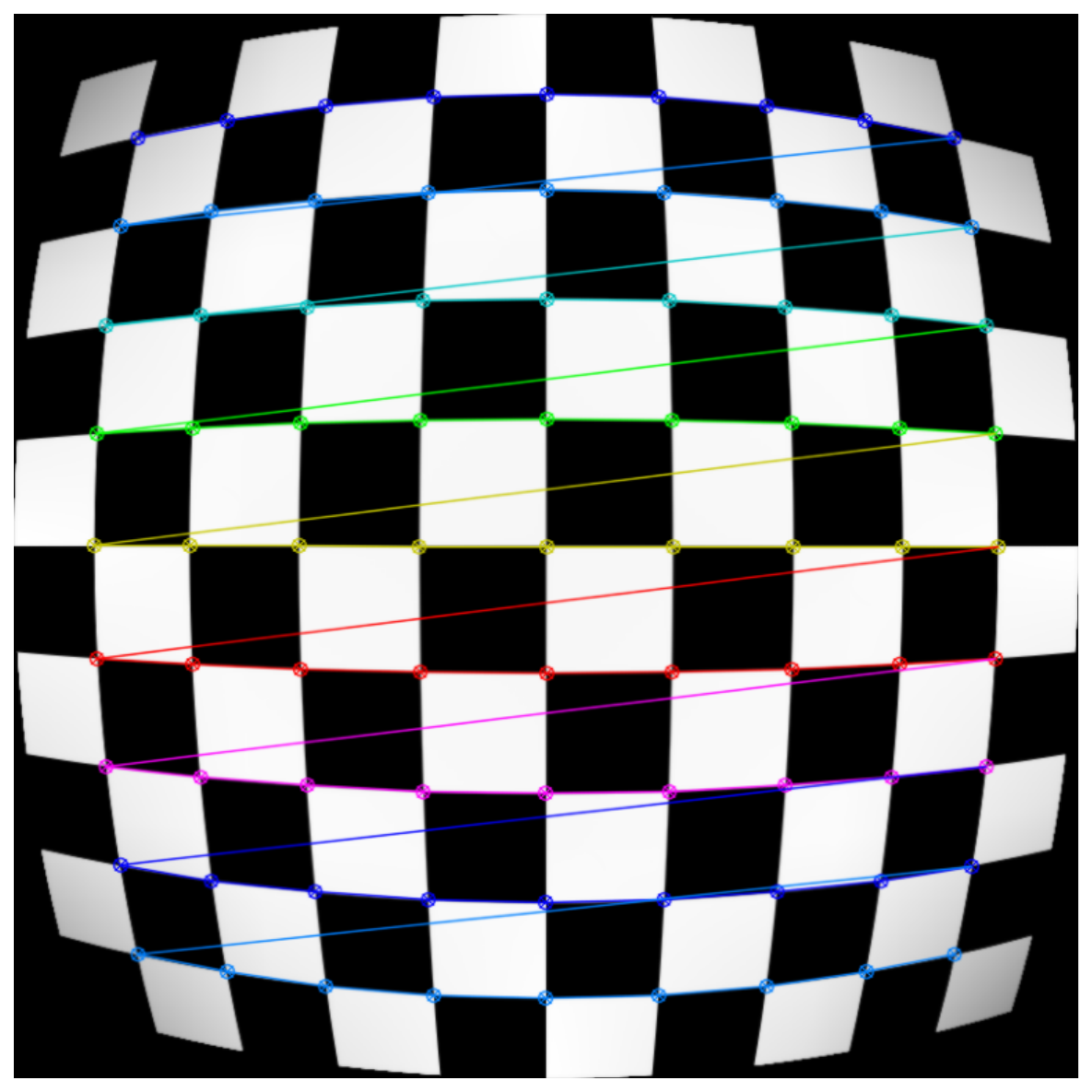}}%
	\hfill
	\subcaptionbox{Code V}{\includegraphics[width=0.15\textwidth]{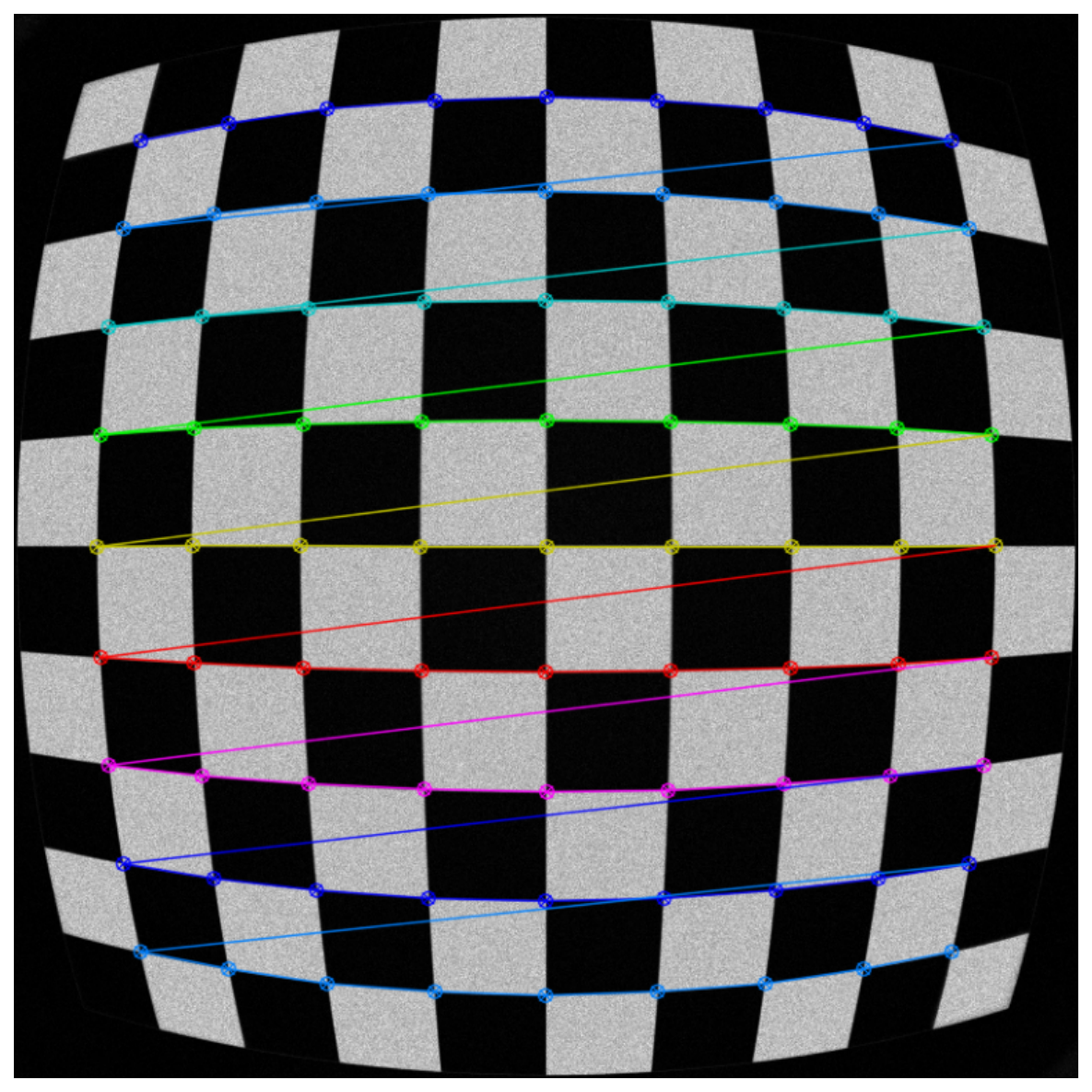}}%
	\hfill
	\subcaptionbox{Corner detection differences}{\includegraphics[width=0.15\textwidth]{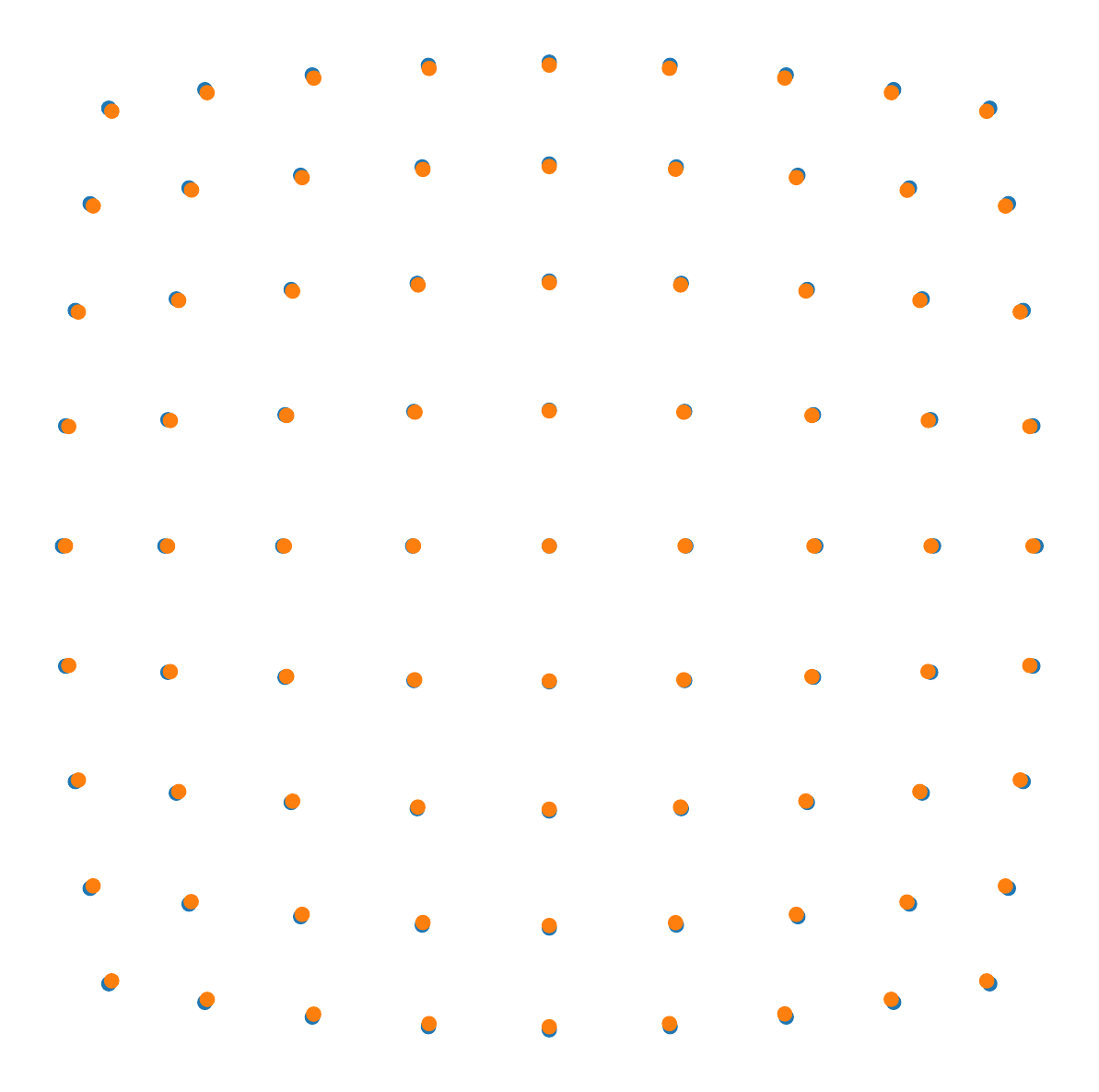}}%
	\caption{Comparison of point matching on a chessboard grid: \package{} (ours) vs. Code V.}\label{fig:chessboard_comparison}
\end{figure}

Simulation accuracy was validated using a chessboard test pattern. Subpixel-accurate corner
positions were extracted from both the Code V and Mitsuba sensor outputs. The mean positional
deviation amounted to approximately 1.2 pixels in both x- and y-directions, indicating close
agreement between design and simulation as can be seen in Figure~\ref{fig:chessboard_comparison}.

\subsection{Comparison to Real Active Alignment Situation}

Established or publicly documented alignment methods for complex multi-lens systems are rare
hindering a realistic and unbiased comparison of different active alignment methods. This is mostly
due to strong confidentiality constraints surrounding industrial active alignment, which typically
cover algorithms, optical designs, hardware, and in some cases even the alignment targets.

\begin{figure}
	\centering
	\subcaptionbox{\package{} (ours)}{\includegraphics[width=0.15\textwidth]{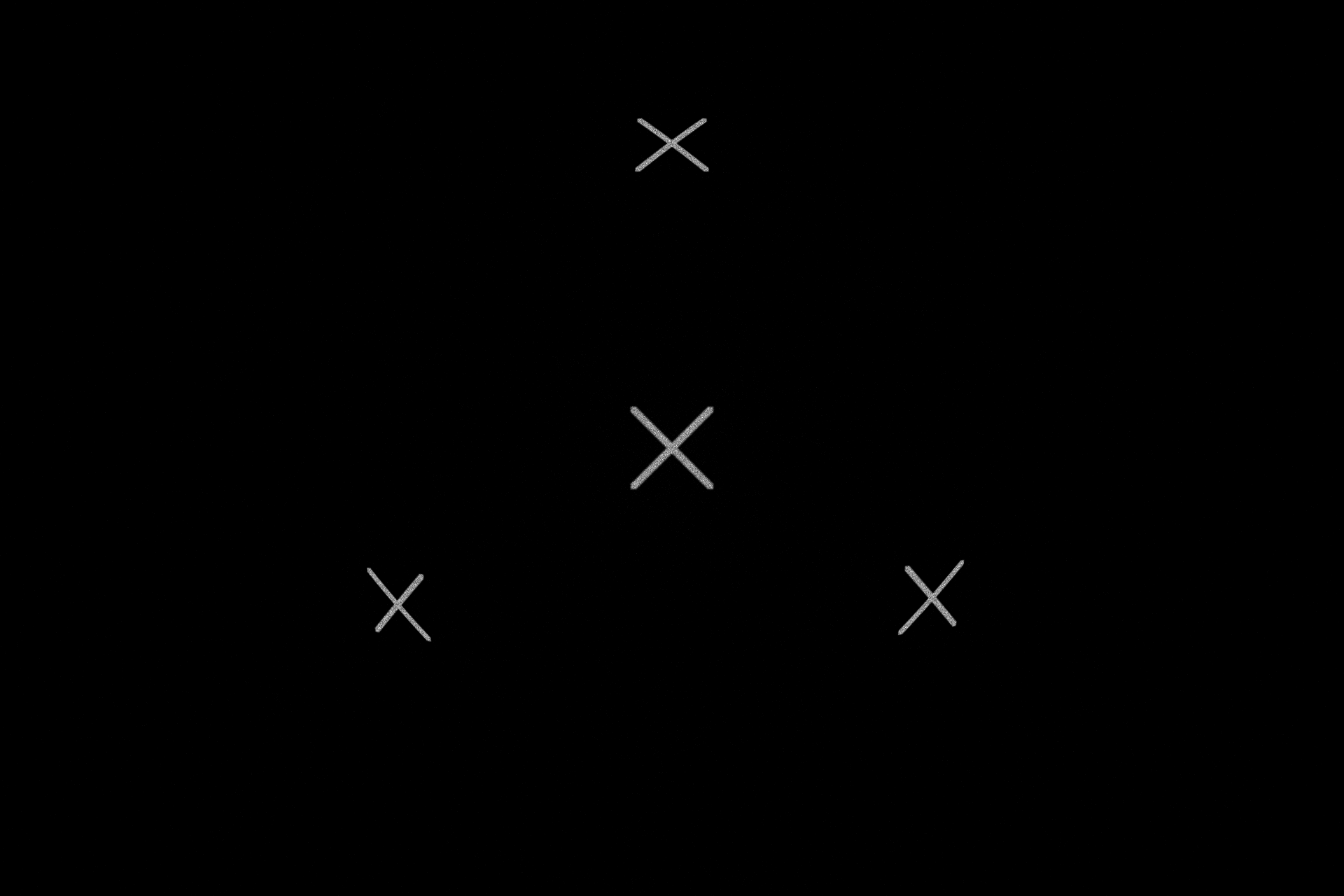}}%
	\hfill
	\subcaptionbox{Real image from production}{\includegraphics[width=0.15\textwidth]{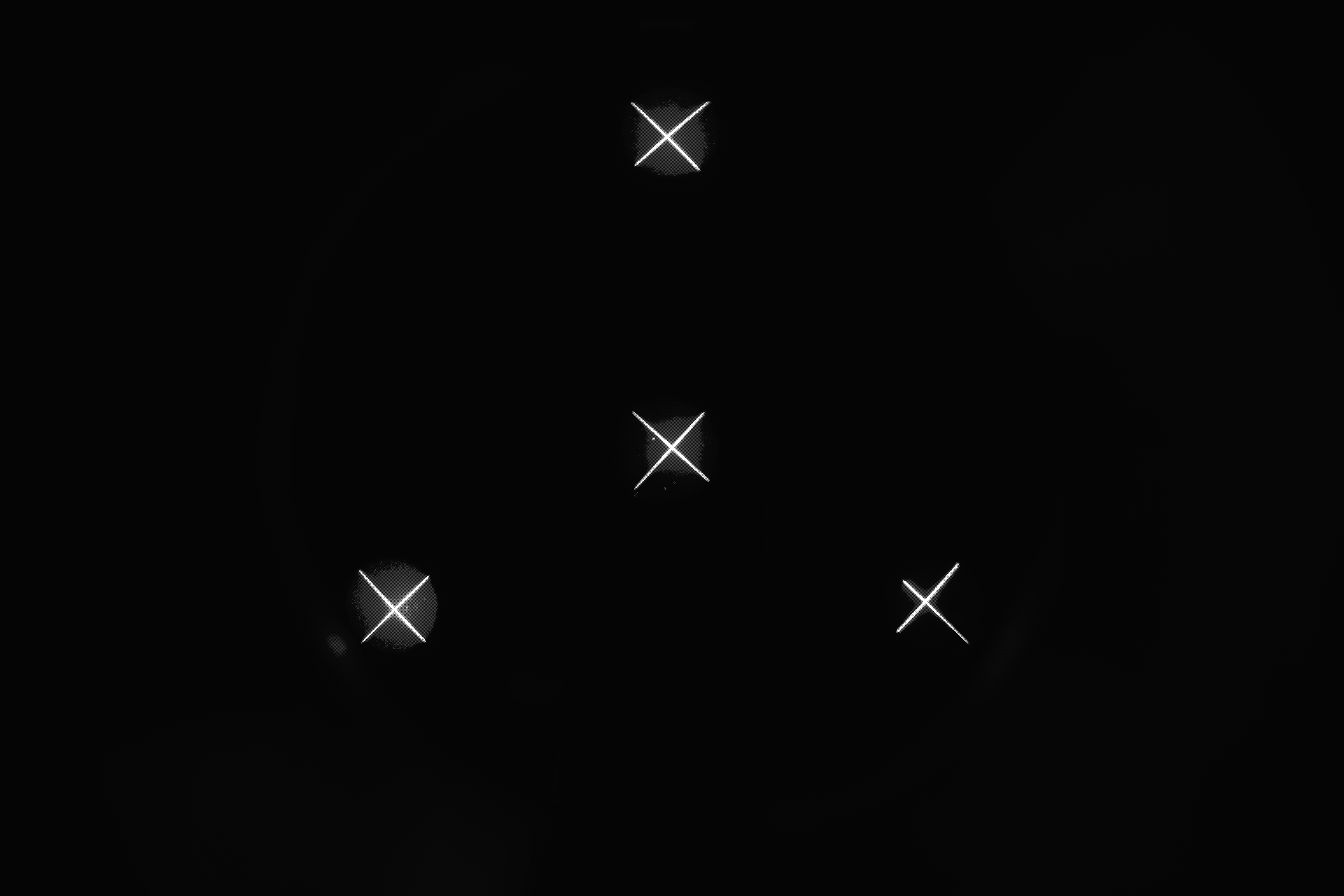}}%
	\hfill
	\subcaptionbox{Difference \package{} (red) and production (green)}{\includegraphics[width=0.15\textwidth]{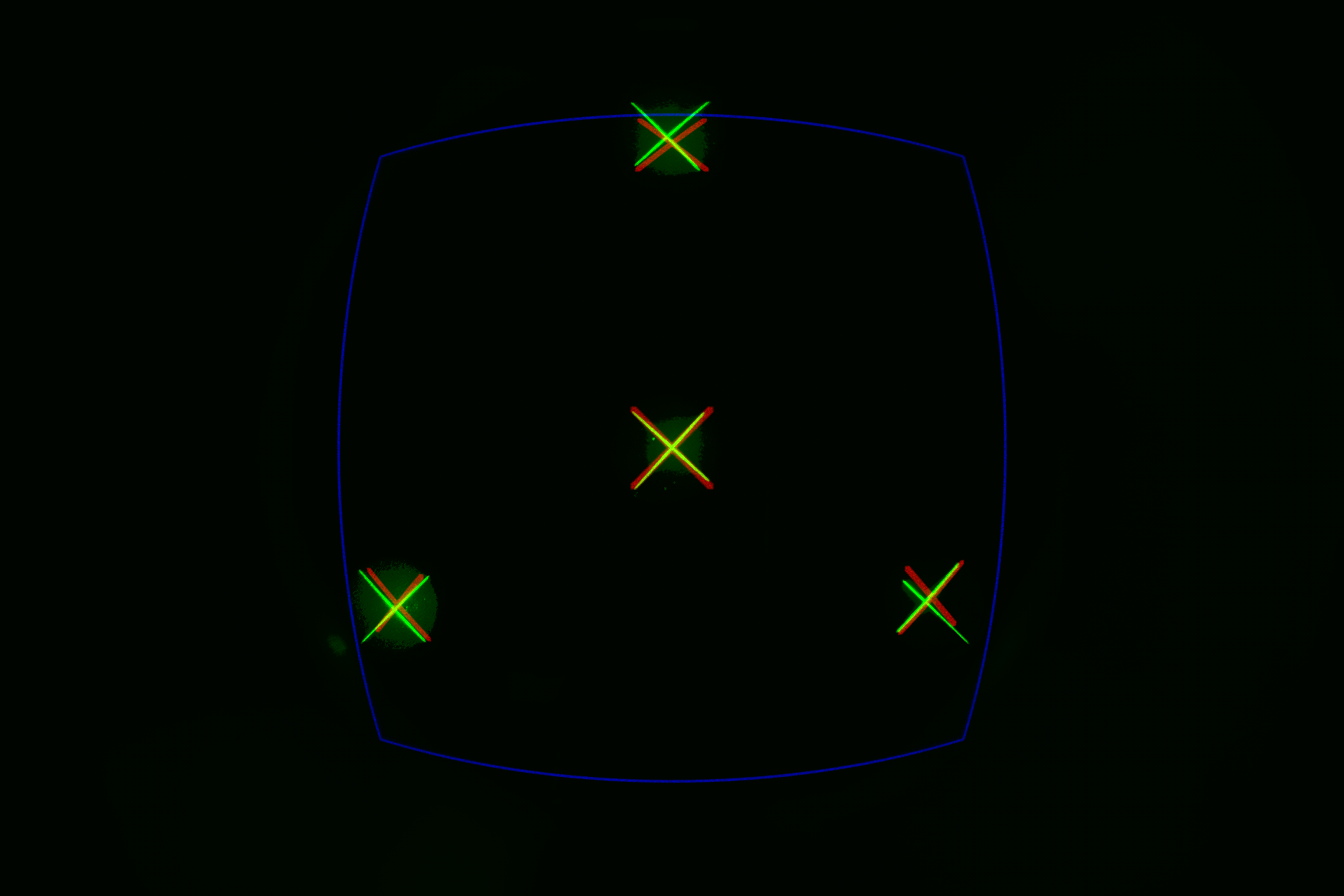}}%
	\caption{Comparison of a realistic alignment target: \package{} (ours) vs. a real alignment
		station.}\label{fig:alignment_production_comparison}

\end{figure}

Consequently, we designed along given the optical parameters of a real-world camera system an
optical design that approximates the performance. We then compare the output of \package{} to a
real sensor output of $1920\times 1280$ pixel obtained through an alignment station. The visual
comparison is shown in Figure~\ref{fig:alignment_production_comparison} demonstrating that
\package{} closely approximates a realistic scenario, with only a small deviation from the
real-world camera.

\section{Experiments}\label{s:experiment}\label{s:results}

We conduct a benchmark of different active alignment methods in this section to showcase how
\package{} can be used. Throughout, we use the optical layout introduced in
Section~\ref{s:optical_layout}.

When rendering images, one typically faces a trade-off between computational efficiency and image
quality. To reduce overall RL training time, we prioritize fast image generation at the cost of
lower pixel resolution and increased sensor variance. Because of that, we only use $200 \times 200$
pixels and $64$ samples per pixel for rendering.

Accurate and comparable benchmarks for active alignments are hard to find, not only due to privacy
constraints of companies maintaining the algorithms and the optical designs as part of their core
know-how, but also due to the fact that optical products are very diverse and have different
requirements. Already small differences in the required optical performance or the assumed
variances of the optical system can lead to completely different alignment strategies and
performances. Typical expert-designed alignment algorithms rely on hand-crafted features extracted
from $O(s, W)$ that need to be optimized simultaneously, often via curve
fitting~\cite{laser_alignment} or expensive area scans. Already for use cases with two degrees of
freedoms, dozens of steps may be required~\cite{laser_alignment}.

A comparison of RL algorithms with classic methods on real systems thus requires a careful design
to make them truly unbiased and insightful and hence this is beyond the scope of this paper.
Nevertheless, we are convinced that our framework eases the effort to design fair and direct
comparison in the future. Instead, we decided to benchmark against state-of-the art optimizers to
solve Equation~(\ref{equ:optimality}) actively and ensure that all algorithms are confronted with
the exact same problem instances.

\subsection{Benchmark Environments}\label{s:environment}

For the evaluation, we focus on three distinct benchmark setups. Each setup considers the same lens
system, but with either none, low or high individual variances $W_L$ for the last two lenses. The
first three components of $W_L$ represent translation offsets along the $x$, $y$ and $z$ axes and
are sampled from a normal distribution $(W_L)_i \sim N_{0, 1.25\cdot 10^{-4}}$ for low and $(W_L)_i
	\sim N_{0, 2.5 \cdot 10^{-4}}$ for high variances with $i \in \{1,2,3\}$. The remaining two
represent rotation offsets along the $x$ and $y$ axes, sampled as $(W_L)_j \sim N_{0, 0.375}$ for
low and $(W_L)_j \sim N_{0, 0.75}$ for high variances with $j \in \{4,5\}$. We denote the
underlying distributions as $\rho_{N}$ for $N=0$, $N=0.25$ and $N=0.5$. These variances account for
potential manufacturing imperfections within a certain tolerance bound. To create a challenging
alignment situation, they are intentionally set very high. Furthermore, rotation around $z$ is
considered redundant, as in a perfectly aligned scenario, rotation around the $z$ axis has no
effect. Note that $O^*$ does not change for setups including object variances.

\subsubsection{RL Algorithms}\label{s:ppo}

\emph{Proximal policy optimization} (PPO) is a policy-based
RL algorithm that trains a stochastic policy
in an on-policy way~\cite{schulman2017proximalpolicyoptimizationalgorithms}.

We used the PPO implementation in Stable Baselines3~\cite{stable-baselines3} and compared it with
other well-tested State-of-the-art RL algorithms. Our analysis revealed PPO's superior performance,
leading us to adopt it as the exclusive RL approach for this study.

To process image observations, we employed a CNN as policy network, that consists of a standard CNN
architecture with three $3\times 3$ convolutional layers, each followed by ReLU activation and
max-pooling. This is followed by a single fully connected layer holding $256$ extracted features
for each image.

As is common in previous work, the learning rate and the entropy coefficient are crucial
hyperpameters of PPO. We experimented with different learning rates in $\{1e^{-2}, 1e^{-3},
	1e^{-4}, 1e^{-5}\}$ and entropy coefficients in $\{1e^{-2}, 1e^{-3}, 1e^{-4}\}$.

We set the discount factor to $\gamma=0.9$ and used the reward function described in
Section~\ref{s:pomdp} with $C=0.1$. The best-performing models were obtained after $1.9\cdot 10^6$
global steps and learning rates of $10^{-3}$ and $10^{-4}$, dependent on $W_L$.

\begin{figure}
	\includegraphics[width=0.48\textwidth]{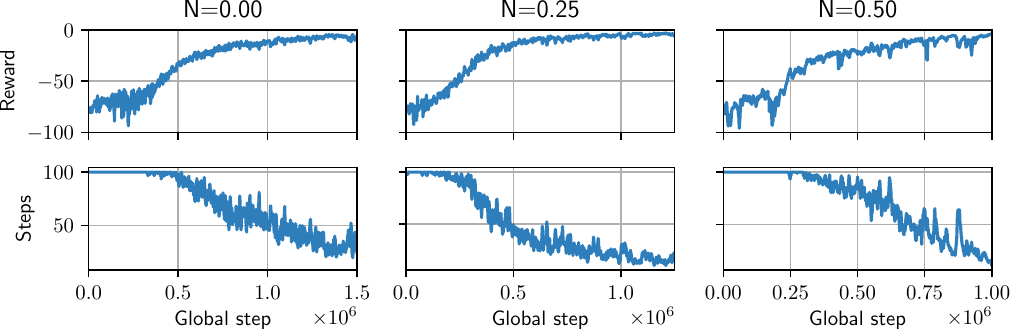}
	\caption{Evolution of rewards during training for different noise levels.}
	\label{fig:reward}
\end{figure}

Figure~\ref{fig:reward} shows the evolution of the reward during online evaluations.
Counterintuitively, one would expect to see a much slower convergence for higher noise during
training. But due to the high variance in high-noise scenarios, the globally defined threshold
$\theta$ must be set at a higher level to ensure that even lower-quality lens system still meet the
acceptance criteria. Otherwise, such lens systems would be classified as not acceptable. However,
increasing the tolerance in this way inevitably leads to a reduction in overall manufacturing
quality.

All trainings are executed on NVIDIA L40S GPUs requiring roughly $0.05$ seconds per global step
resulting in a total train time of approximately $28$ hours. The full details can be found in the
accompanying code of this paper.

\subsubsection{Bayesian Optimization}\label{s:bo}

Bayesian Optimization (BO) is a strategy for minimizing functions $f$ that are expensive to
evaluate by building a probabilistic model of the objective function~\cite{bo,smbo}. It selects new
evaluation points by balancing exploration and exploitation using an acquisition function, which
predicts where the function is likely to improve. This approach is particularly useful when
function evaluations are costly, as it finds optimal solutions with relatively few evaluations.
Setting $f_W(s):=-R(s, W)$, an alignment problem can be interpreted as an optimization problem for
the family of black box functions $f_W$. We evaluated BO algorithms using a different probabilistic
model than implemented in scikit-optimize~\cite{scikit-optimize}: Gaussian Processes
(BO-GP)~\cite{bo_gp} and Random Forests (BO-RF)~\cite{bo_forrests}. As vanilla BO algorithms
explore~$f_W$ for each $W$ and without a priori information, the state space has to be explored
first randomly costing unnecessary steps. Thus, we also tested the method proposed
in~\cite{transfergp} (TransferGP) that allows pre-training a Gaussian process on samples from
problem instances $f_{W_1},\ldots,f_{W_m}$.

\subsubsection{Random}

As a baseline, we implemented an algorithm that samples uniformly at random from the alignment
space.

\subsection{Results}

\begin{figure}
	\centering
	\includegraphics[width=0.45\textwidth]{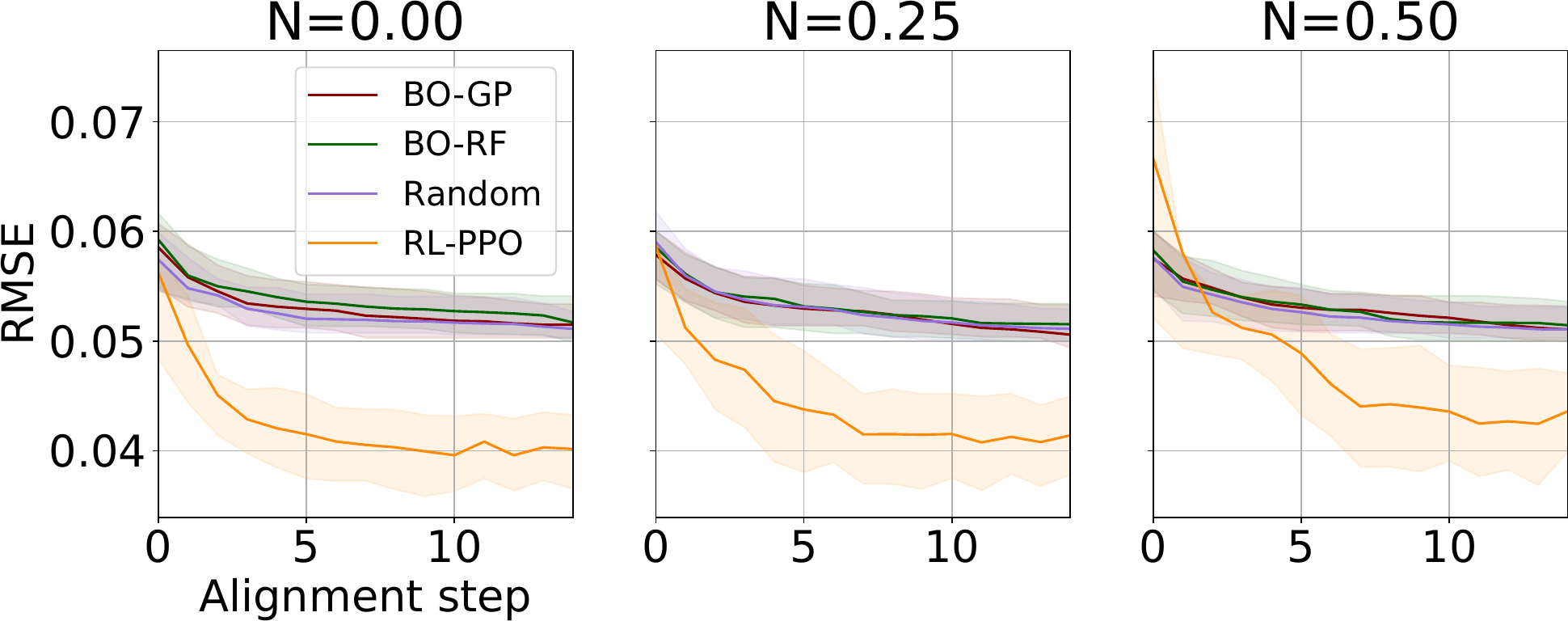}
	\caption{Comparison of baseline alignment algorithms against RL-PPO for different noise levels.}
	\label{fig:baseline}
\end{figure}

For evaluation, each algorithm was executed on $100$ different environments for each of the
benchmark situation described in Section~\ref{s:environment}. We compared approaches that operate
without any a priori information and require no domain knowledge. Except for the RL algorithms, the
algorithms used do not involve a training phase. To address this imbalance, we reduced the search
space for baseline algorithms to approximately eight percent of the search space used for RL. The
root mean squared error (RMSE), computed over all pixels, was used as the performance metric.
Figure~\ref{fig:baseline} shows that the RL-based method surpasses in all scenarios all other
algorithms in terms of convergence speed. Moreover, independent of the noise level, RL-PPO reaches
its optimum in under ten steps. As expected, the benchmarks with larger noise levels converge at a
higher error, because due to the lens variances, even the best alignment cannot reach the minimal
render variance error.

Only considering the computation time, RL-PPO requires a constant 25 ms and BO-RF 95 ms per
alignment step. In contrast, BO-GP becomes increasingly time-consuming as the number of steps
increases. For 20 steps, each step takes an average of 100 ms, while for 50 steps the time per step
increases to 178 ms. For the pre-trained Gaussian processes of TransferGP, the processing time for
each step increases with the number of instances it has been pre-trained on. Already when trained
on $m=10$ instances with $100$ samples each, their processing time per step takes several minutes
while their performance equals almost the performance of vanilla BO-GP. Due to their impractical
computation time, we have not included pre-trained GP models in our benchmark.

\begin{figure}
	\centering
	\includegraphics[width=0.45\textwidth]{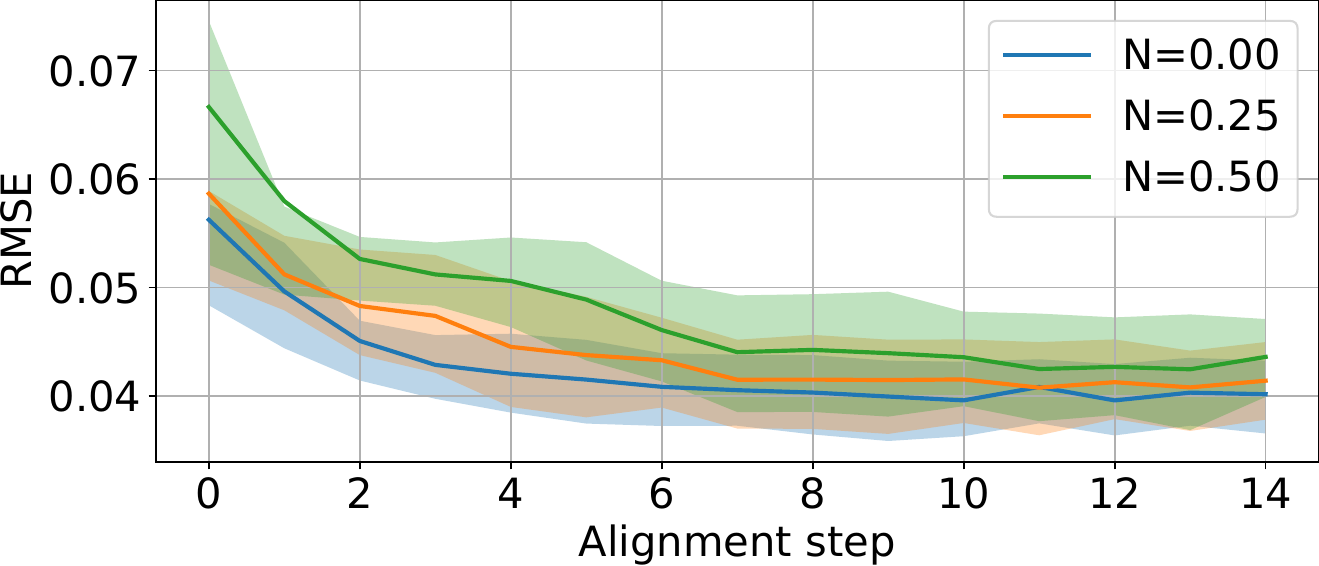}
	\caption{Evaluation of RMSE for the best PPO models and different noise levels.}
\end{figure}

\section{Conclusion}

This work introduced an RL approach for active alignments of optical components. Unlike traditional
alignment methods that rely on expert-designed alignment concepts involving the computation of
hand-crafted features, our approach learns optimal alignment strategies directly from
high-dimensional sensor observations. By leveraging RL, we demonstrated that alignment tasks can be
solved more efficiently, even in the presence of noise and manufacturing tolerances. However, the
low inference time of RL-algorithms at runtime comes at the price of many training iterations. Our
experiments show that RL-based alignment not only outperforms conventional machine learning
approaches in terms of efficiency but also eliminates the need for manually designed features. This
work opens the door for further exploration of RL in high-precision optical assembly, with
potential applications in automated manufacturing, adaptive optics, and real-time calibration of
complex optical systems. Future research could focus on improving sample efficiency, integrating
domain adaptation techniques, and extending RL-based alignment to real-world hardware
implementations.


\newpage
\bibliographystyle{ieeetr}
\bibliography{relign.bib}

\end{document}

%% file: figures/overview.tex
\begin{tikzpicture}

	\node[inner sep=2] (distorted) at (0, 0) {
		\includegraphics[width=2cm]{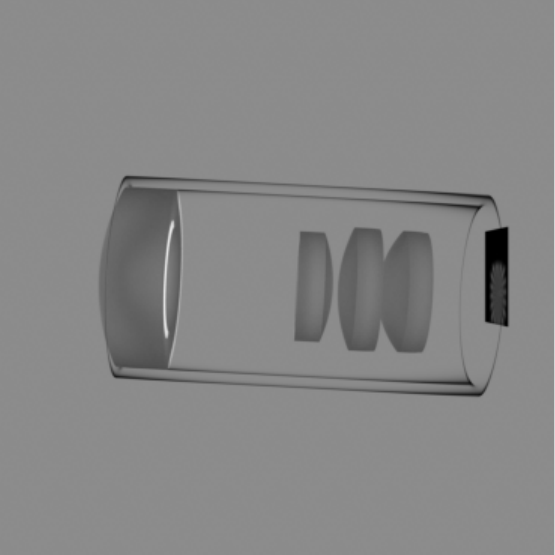}
	};

	\node[inner sep=2] (distorted_sensor) at (4, 0) {
		\includegraphics[width=1.5cm]{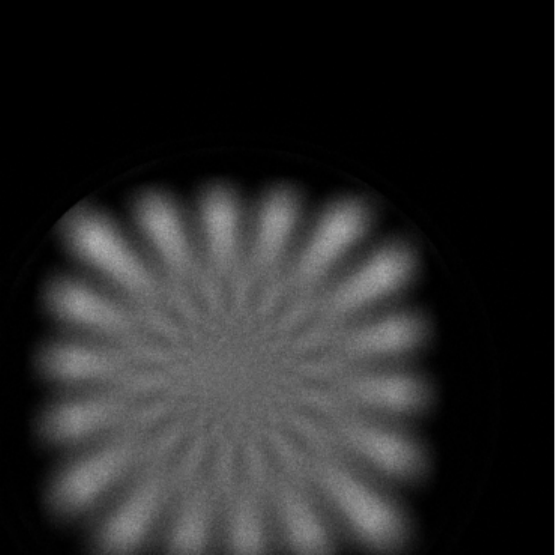}
	};
	\node[inner sep=2] (aligned) at (0, -3) {
		\includegraphics[width=2cm]{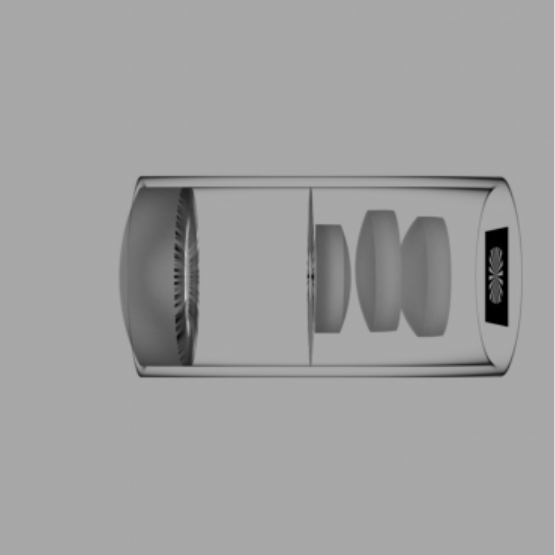}
	};
	\node[inner sep=1] (aligned_sensor) at (4, -3) {
		\includegraphics[width=1.5cm]{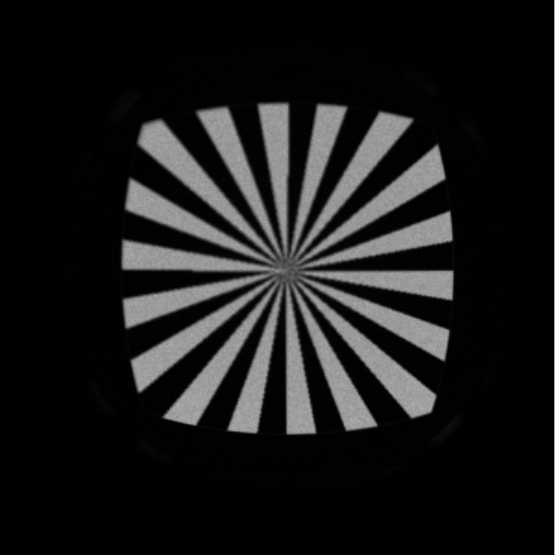}
	};

	\draw[-latex, very thick, dotted] (distorted) to (distorted_sensor);
	\draw[-latex, very thick, dotted] (aligned) to (aligned_sensor);

	\draw[-latex, very thick] (3, -0.7) to[bend right]
	node[midway, fill=white,scale=1.5] {$\pi$} (1.0, -1.9);

	\node[left=2.2cm of distorted, scale=1.5, anchor=west] {$(s_i, W)$};
	\node[left=2.2cm of aligned, scale=1.5, anchor=west] {$(s_{i+1}, W)$};
	\node[right=0.2cm of distorted_sensor, scale=1.5, anchor=west] {$O(s_i, W)$};
	\node[right=0.2cm of aligned_sensor, scale=1.5, anchor=west] {$O(s_{i+1}, W)$};

\end{tikzpicture}

%% file: relign.bbl
\begin{thebibliography}{10}

\bibitem{yoder2008mounting}
P.~Yoder, {\em Mounting Optics in Optical Instruments}.
\newblock Online access with subscription: SPIE Digital Library, SPIE, 2008.

\bibitem{alignment_strategies}
P.~Langehanenberg, J.~Heinisch, C.~Wilde, F.~Hahne, and B.~L{\"u}er{\ss},
  ``{Strategies for active alignment of lenses},'' in {\em Optifab 2015} (J.~L.
  Bentley and S.~Stoebenau, eds.), vol.~9633, p.~963314, International Society
  for Optics and Photonics, SPIE, 2015.

\bibitem{RobotAssemblyLargeScaleObjects}
Z.~Qin, P.~Wang, J.~Sun, J.~Lu, and H.~Qiao, ``Precise robotic assembly for
  large-scale objects based on automatic guidance and alignment,'' {\em IEEE
  Transactions on Instrumentation and Measurement}, vol.~65, no.~6,
  pp.~1398--1411, 2016.

\bibitem{PoseEstimationLargeScaleObjects}
S.~Liu, D.~Xu, F.~Liu, D.~Zhang, and Z.~Zhang, ``Relative pose estimation for
  alignment of long cylindrical components based on microscopic vision,'' {\em
  IEEE/ASME Transactions on Mechatronics}, vol.~21, no.~3, pp.~1388--1398,
  2016.

\bibitem{FastAlignmentPhotonicsDevice}
J.~Guo and R.~Heyler, ``Fast active alignment in photonics device packaging,''
  in {\em 2004 Proceedings. 54th Electronic Components and Technology
  Conference (IEEE Cat. No.04CH37546)}, vol.~1, pp.~813--817 Vol.1, 2004.

\bibitem{offline_alignment}
D.~Zontar, A.~Tavakolian, M.~Hoeren, and C.~Brecher, ``{Offline development of
  active alignment based on empirical virtual environments},'' in {\em
  High-Power Diode Laser Technology XVIII} (M.~S. Zediker, ed.), vol.~11262,
  p.~112620B, International Society for Optics and Photonics, SPIE, 2020.

\bibitem{rakhmatulin2024review}
I.~Rakhmatulin, D.~Risbridger, R.~M. Carter, M.~D. Esser, and M.~S. Erden, ``A
  review of automation of laser optics alignment with a focus on machine
  learning applications,'' {\em Optics and Lasers in Engineering}, vol.~173,
  p.~107923, 2024.

\bibitem{supervised_alignments}
K.~Hinrichs and J.~Piotrowski, ``Neural networks for faster optical
  alignment,'' {\em Optical Engineering}, vol.~59, pp.~074107 -- 074107, 2020.

\bibitem{ml_for_off_axis_systems}
L.~Yu, C.~Ma, X.~Fu, Y.~Yin, and M.~Cao, ``{Application of machine learning in
  the alignment of off-axis optical system},'' in {\em AOPC 2021: Novel
  Technologies and Instruments for Astronomical Multi-Band Observations}
  (Y.~Zhu and S.~Xue, eds.), vol.~12069, p.~120690P, International Society for
  Optics and Photonics, SPIE, 2021.

\bibitem{ModelBasedAlignment}
M.~Hoeren, D.~Zontar, A.~Tavakolian, M.~Berger, S.~Ehret, T.~Mussagaliyev, and
  C.~Brecher, ``{Performance comparison between model-based and machine
  learning approaches for the automated active alignment of FAC-lenses},'' in
  {\em High-Power Diode Laser Technology XVIII} (M.~S. Zediker, ed.),
  vol.~11262, p.~1126209, International Society for Optics and Photonics, SPIE,
  2020.

\bibitem{alphago}
D.~Silver, A.~Huang, C.~J. Maddison, A.~Guez, L.~Sifre, G.~Van Den~Driessche,
  J.~Schrittwieser, I.~Antonoglou, V.~Panneershelvam, M.~Lanctot, {\em et~al.},
  ``Mastering the game of go with deep neural networks and tree search,'' {\em
  Nature}, vol.~529, no.~7587, pp.~484--489, 2016.

\bibitem{dreamer}
D.~Hafner, T.~Lillicrap, J.~Ba, and M.~Norouzi, ``Dream to control: Learning
  behaviors by latent imagination,'' in {\em International Conference on
  Learning Representations}, 2020.

\bibitem{alphafold}
J.~M. Jumper, R.~Evans, A.~Pritzel, T.~Green, M.~Figurnov, O.~Ronneberger,
  K.~Tunyasuvunakool, R.~Bates, A.~Ž{\'i}dek, A.~Potapenko, A.~Bridgland,
  C.~Meyer, S.~A.~A. Kohl, A.~Ballard, A.~Cowie, B.~Romera-Paredes, S.~Nikolov,
  R.~Jain, J.~Adler, T.~Back, S.~Petersen, D.~Reiman, E.~Clancy, M.~Zielinski,
  M.~Steinegger, M.~Pacholska, T.~Berghammer, S.~Bodenstein, D.~Silver,
  O.~Vinyals, A.~W. Senior, K.~Kavukcuoglu, P.~Kohli, and D.~Hassabis, ``Highly
  accurate protein structure prediction with alphafold,'' {\em Nature},
  vol.~596, pp.~583 -- 589, 2021.

\bibitem{rl_for_process_control}
R.~Nian, J.~Liu, and B.~Huang, ``A review on reinforcement learning:
  Introduction and applications in industrial process control,'' {\em Computers
  \& Chemical Engineering}, vol.~139, p.~106886, 2020.

\bibitem{rl_for_laser_optics}
I.~Rakhmatulin, D.~Risbridger, R.~M. Carter, M.~J.~D. Esser, and M.~S. Erden,
  ``Reinforcement learning for aligning laser optics with kinematic mounts,''
  {\em 2024 IEEE 20th International Conference on Automation Science and
  Engineering (CASE)}, pp.~1397--1402, 2024.

\bibitem{sorokin2020interferobot}
D.~Sorokin, A.~Ulanov, E.~Sazhina, and A.~Lvovsky, ``Interferobot: aligning an
  optical interferometer by a reinforcement learning agent,'' {\em Advances in
  Neural Information Processing Systems}, vol.~33, pp.~13238--13248, 2020.

\bibitem{rl_challenges}
C.~Paduraru, D.~Mankowitz, G.~Dulac-Arnold, J.~Li, N.~Levine, S.~Gowal, and
  T.~Hester, ``Challenges of real-world reinforcement learning: definitions,
  benchmarks and analysis,'' {\em Machine Learning}, vol.~110, pp.~2419 --
  2468, 2021.

\bibitem{delayed_reward}
S.~F. Chevtchenko and T.~B. Ludermir, ``Learning from sparse and delayed
  rewards with a multilayer spiking neural network,'' in {\em 2020
  International Joint Conference on Neural Networks (IJCNN)}, pp.~1--8, 2020.

\bibitem{rl_efficiency}
S.~Kamthe and M.~Deisenroth, ``Data-efficient reinforcement learning with
  probabilistic model predictive control,'' in {\em Proceedings of the
  Twenty-First International Conference on Artificial Intelligence and
  Statistics} (A.~Storkey and F.~Perez-Cruz, eds.), vol.~84 of {\em Proceedings
  of Machine Learning Research}, pp.~1701--1710, PMLR, 09--11 Apr 2018.

\bibitem{rl_reproducability}
N.~A. Lynnerup, L.~Nolling, R.~Hasle, and J.~Hallam, ``A survey on
  reproducibility by evaluating deep reinforcement learning algorithms on
  real-world robots,'' in {\em Proceedings of the Conference on Robot Learning}
  (L.~P. Kaelbling, D.~Kragic, and K.~Sugiura, eds.), vol.~100 of {\em
  Proceedings of Machine Learning Research}, pp.~466--489, PMLR, 30 Oct--01 Nov
  2020.

\bibitem{Mitsuba3}
W.~Jakob, S.~Speierer, N.~Roussel, M.~Nimier-David, D.~Vicini, T.~Zeltner,
  B.~Nicolet, M.~Crespo, V.~Leroy, and Z.~Zhang, ``Mitsuba 3 renderer,'' 2022.
\newblock https://mitsuba-renderer.org.

\bibitem{gymnasium}
M.~Towers, A.~Kwiatkowski, J.~Terry, J.~U. Balis, G.~De~Cola, T.~Deleu,
  M.~Goul{\~a}o, A.~Kallinteris, M.~Krimmel, A.~KG, {\em et~al.}, ``Gymnasium:
  A standard interface for reinforcement learning environments,'' {\em arXiv
  preprint arXiv:2407.17032}, 2024.

\bibitem{siemens_star}
G.~C. Birch and J.~C. Griffin, ``{Sinusoidal Siemens star spatial frequency
  response measurement errors due to misidentified target centers},'' {\em
  Optical Engineering}, vol.~54, no.~7, p.~074104, 2015.

\bibitem{iso}
``{Digital cameras — Resolution and spatial frequency responses},'' {2024}.

\bibitem{contextual_mdp}
D.~Ghosh, J.~Rahme, A.~Kumar, A.~Zhang, R.~P. Adams, and S.~Levine, ``Why
  generalization in {RL} is difficult: Epistemic {POMDP}s and implicit partial
  observability,'' in {\em Advances in Neural Information Processing Systems}
  (A.~Beygelzimer, Y.~Dauphin, P.~Liang, and J.~W. Vaughan, eds.), 2021.

\bibitem{Synopsys_CodeV_2024}
I.~Synopsys, ``Code v.'' Proprietary software, version 2024.03, 2024.

\bibitem{laser_alignment}
F.~Liu, Z.~Liu, X.~Li, and D.~Jiang, ``Laser active alignment algorithm based
  on spot features and curve fitting,'' in {\em 2023 8th International
  Conference on Automation, Control and Robotics Engineering (CACRE)},
  pp.~150--155, 2023.

\bibitem{schulman2017proximalpolicyoptimizationalgorithms}
J.~Schulman, F.~Wolski, P.~Dhariwal, A.~Radford, and O.~Klimov, ``Proximal
  policy optimization algorithms,'' 2017.

\bibitem{stable-baselines3}
A.~Raffin, A.~Hill, A.~Gleave, A.~Kanervisto, M.~Ernestus, and N.~Dormann,
  ``Stable-baselines3: Reliable reinforcement learning implementations,'' {\em
  Journal of Machine Learning Research}, vol.~22, no.~268, pp.~1--8, 2021.

\bibitem{bo}
X.~Wang, Y.~Jin, S.~Schmitt, and M.~Olhofer, ``Recent advances in bayesian
  optimization,'' {\em ACM Computing Surveys}, vol.~55, pp.~1 -- 36, 2022.

\bibitem{smbo}
Q.~Zhang and Y.~Hwang, ``Sequential model-based optimization for continuous
  inputs with finite decision space,'' {\em Technometrics}, vol.~62, pp.~486 --
  498, 2019.

\bibitem{scikit-optimize}
T.~scikit-optimize contributors, ``scikit-optimize: Sequential model-based
  optimization in python,'' 2024.

\bibitem{bo_gp}
C.~Rasmussen and C.~Williams, {\em Gaussian Processes for Machine Learning}.
\newblock Adaptive Computation and Machine Learning, Cambridge, MA, USA: MIT
  Press, Jan. 2006.

\bibitem{bo_forrests}
A.~Lacoste, H.~Larochelle, M.~Marchand, and F.~Laviolette, ``Sequential
  model-based ensemble optimization.,'' in {\em UAI} (N.~L. Zhang and J.~Tian,
  eds.), pp.~440--448, AUAI Press, 2014.

\bibitem{transfergp}
P.~Tighineanu, K.~Skubch, P.~Baireuther, A.~Reiss, F.~Berkenkamp, and
  J.~Vinogradska, ``Transfer learning with gaussian processes for bayesian
  optimization,'' in {\em Proceedings of The 25th International Conference on
  Artificial Intelligence and Statistics} (G.~Camps-Valls, F.~J.~R. Ruiz, and
  I.~Valera, eds.), vol.~151 of {\em Proceedings of Machine Learning Research},
  pp.~6152--6181, PMLR, 28--30 Mar 2022.

\end{thebibliography}
